# From melodic note sequences to pitches using word2vec

Daniel Defays[1]

[1]Université of Liège, Department of Psychology, ddefays@uliege.be

Applying the word2vec technique, commonly used in language modeling, to melodies—where notes are treated as words in sentences—enables the capture of pitch information. This study examines two datasets: 20 children's songs and an excerpt from a Bach sonata. The semantic space for defining the embeddings is of very small dimension, specifically 2. Notes are predicted based on the 2, 3 or 4 preceding notes that establish the context. A multivariate analysis of the results shows that the semantic vectors representing the notes have a multiple correlation coefficient of approximately 0.80 with their pitches.

**Keywords**

Embedding; Machine Learning; Semantic meaning; Correlation

## 1. Introduction

What kind of meaning can we capture from musical notes using word embedding techniques typically applied in language models? This study addresses this question by modeling various types of music with a relatively simple neural network, commonly used for word embedding.
An embedding is a vector representation of an entity (a word, an image, a sound) in a multidimensional space where geometric relationships between vectors reflect semantic relationships between the corresponding entities (Chollet, 2021). This inquiry is not new; numerous statistical and computational models, including neural networks, have been proposed to capture key features of musical pieces and to model music perception. In 2016, Madjiheurem, Qu and Walder compared different embedding techniques to learn musical chord embeddings. Their analysis used chords as entry and output, and high dimensional spaces to represent chords. Herremans and Chuan (2017) used a slightly different approach. They cut pieces of classical music into slices and used, as Madjiheurem, Qu and Walder (2016), word2vec (Mikolov, 2013), an efficient approach to associate vectors to words (here musical slices) in natural language processing. This method captured features such as the tonal proximity of slices. Korzeniowski, in 2018, proposed to use other types of codification of the musical pieces and another deep learning model. His approach involved processing the audio signal with a Fourier transform before feeding it into a convolutional neural network to extract descriptors of its harmonic content. Garcia-Valencia in 2020, used a recurrent neural network based on LSTM to represent musical meaning of notes. Three different datasets were used to test the model, with input in the form of binary one-hot vectors and a vector space of 128 dimensions. His model was designed to predict the subsequent note from each given note, enabling the visualization of meaningful groupings of notes based on their relative position in their octave or the presence of alterations. Hongru Liang, Wenqiang Lei, Paul Yaozhu Chan, Zhenglu Yang, Maosong Sun and Tat-Seng Chua (2020) proposed to integrate pitch, rhythm and dynamics plus information on the melodic context in the embedding approach. This enables among others to capture musical information relevant for music completion or genre classification of songs. To model the early stages of musical perception, Defays, French, and Tillmann (2023) demonstrated that TRACX2, a recognition-based, recursive connectionist autoencoder model of chunking and sequence segmentation, can reproduce some features of human musical perception. They started with intervals between notes, measured in semitones, and replaced the traditional one-hot encoding scheme with what they called an ordinal encoding scheme, another method for encoding musical notes. Other algorithms have been developed in the field of music-information retrieval (MIR) to automatically detect and extract repeated patterns from musical scores (Müller & Clausen, 2007; Nieto & Farbood, 2014) or sound files.

My approach diverges from previous work in several significant ways. I employ a particularly simple neural model, making it easy to understand. This model is designed to predict an upcoming note based on the preceding ones, utilizing the CBOW approach (T. Mikolov, K. Chen, G. Corrado and J. Dean, 2013), rather than the traditionally used Skip-gram model for musical embeddings. The representations obtained utilize very low-dimensional spaces, making them easy to interpret without the need for projection techniques such as t-SNE or PCA. The music used in the study spans different genres, including children's nursery rhymes and a Bach sonata. To my knowledge, the results are quite distinct from what has been observed so far. They establish a strong link between the pitches of the notes and their sequential order in a melody.

## 1.1 Objective of the study

The recent breakthroughs in machine learning and more specifically in natural language processing are partly explained by the way words are represented. Most language models start from a numerical representation of words (or part of words called tokens) in so called semantic spaces. The dimension of those spaces generally varies from 128 to several thousands. The semantic representation of words, also called embeddings, is obtained with simple neural network designed to predict missing words in sentences. The input data – sequence of words or tokens – are fed with one-hot vectors (see below for an explanation of this type of coding) into the network, they flow through a hidden layer whose dimension defines the dimension of the semantic space and produce an output vector supposed to approach the one-hot vector defining the target word to predict. The words are so mapped by the neural network to vectors of real numbers called embeddings that carry out part of their « meaning ». Vectors representing words that are similar in a semantic sense (like "song" and "melody") have a smaller distance between them than vectors representing words that have no semantic relationship.

What happens if such a procedure is applied to the sequence of musical notes constituting a melody? This is the topic of this study. It seems interesting to see how far one can go without an « intelligent » coding of the data as proposed in the paper of Defays, French and Tillmann (2023) for instance. Given that in such a procedure the model does not know the pitches of the notes how far can it go in their representation? Will we find a link with some musical properties of the notes? This is the question I address.

## 1.2 Musical notes encoding

Most approaches use to feed the data into the model a simple encoding scheme called one-hot. This encoding scheme is traditionally used in language models to represent sentences as a sequence of words or tokens. The principle is easy to understand. Let us assume our melody starts with the following sequence of musical notes: E4 E4 C4 E4 G4 C4 C4 D4 F4 D4 D4 E4 C4 … To encode the data, you need to create first a vocabulary with all the different notes used in the melody. Let us assume that the size of the vocabulary is 16; this means the song requires 16 different notes. Each note is then represented by a vector of size 16, where each component is set to 0, except one that is set to 1. This only 1 is in the position of the corresponding note in the vocabulary. In our example, if the vocabulary is V = (C4, D4, E4, G4 …), C4 will be coded as (1,0,0,0, …), D4 as (0,1,0,0, …), E4 as (0,0,1,0, …) and so on.

To represent more than one note, one can either use different one-hot vectors, one for each note, or use a multi-hot vector. In a multi-hot encoding, the components of the vector that correspond to the notes are set to 1 and all the other ones are set to 0. For instance, the pair of notes {D4, G4} will be coded (0,1,0,1,0 …).

Defays, French and Tillmann (2023), in their exploration, were forced, in order to reproduce human like feature of music perception as mentioned above, to replace the traditional one-hot encoding scheme by what they called an ordinal encoding scheme. In an ordinal encoding, the order relationship between the different pitches (or the different intervals) is taken into consideration.

## 2. Material and methods

### 2.1 General procedure

The simplest form of embedding network is used in this study. It is applied to sequences of notes in a melody to predict the up-coming element. The embedding generated by the network is then studied from a statistical viewpoint.

### 2.2 Input data

I use as datasets the melodies studied by D. Defays, R. French and B. Tillmann (2023) in their exploration on musical perception. The first set is a set of 20 French children's songs in which only pitches are considered, all with equal duration. The notes are encoded by their chromatic degree (12 values per octave, consequently) without reducing them to a single octave via a modulo 12 transformation, as is often done in segmentation analyses. The chosen key is the one given in the score and all the songs are in the major mode. Features, such as rhythm, meter, tempo, harmony, and texture are not considered. These songs are: *Ah les crocodiles; Bateau sur l'eau; Fais dodo, Colas mon p'tit frère; Au clair de la lune ; Ainsi font; Une souris verte; Ah vous dirai-je maman; Pomme de reinette; Sur le pont d'Avignon; Frappe, frappe, petite main ; Alouette, gentille alouette; Biquette ne veut pas sortir du chou; Dans la forêt lointaine; Je te tiens, tu me tiens; Le bon roi Dagobert; Il était une bergère; J'ai du bon tabac; J'ai perdu le do; Frère Jacques; Il court le furet*. The second set is used to see how far the results can generalize to other kind of music. It is defined by the first 42 measures of the Allegro Assai of the Sonata for Violin Solo in C Major BWV 1005 by J. S. Bach. The Sonata is encoded similarly to the songs: as a sequence of pitches without information on duration, rhythm, and other features.

### 2.3 Neural network

The neural network is a simple feed forward network with the following key features:

- it is a three-layer network designed to predict the up-coming musical note (the target) in a melody. It starts from its context, the preceding notes. The context size varies from 1 to 4 in this study; an example of the prediction task is given below;

- the input data fed into the model are one-hot or multi-hot vectors depending on the size of the context. This means the size of the input depends on the vocabulary (16 different notes for the children's songs and 25 for the Sonata);

- the model is trained to generate prediction of the target. This makes it possible to adjust the weights of the different matrices implied in the computation. The lost function is the traditional cross-entropy. The model is thus a CBOW (Continuous Bag-Of-Words) model;

- the middle layer, that defines the embedding, has two nodes. In traditional networks designed to define word embedding, the size of that layer is much larger;

- there is no activation function attached to the nodes of the second layer (this means a linear activation function), and the last layer is a softmax layer.

The meta-parameters of the network used are as follows:

- Learning rate: 0.01
- Epochs: 40
- Percentage of data used for training: 90% (the rest is used for model validation)

## 2.4 Processing of the data

The input data, after transformation, are subdivided into training data and validation data. This means, for the training data, 843 couples of notes for the children's songs with a context size of 2, and 437 for the Sonata.

As already explained, the data are then processed by the neural network whose weights are progressively adjusted to estimate the target note. To make it perfectly clear, for the sequence of notes I gave above, the model with a context size of 2, for example, must predict

- C4 from {E4, E4},
- E 4 from {E4, C4},
- F4 from {C4, E4}, and so on.

Two important remarks must be made regarding the way the input is processed. Firstly, the value of the notes, their pitch (here their chromatic degree), is completely ignored in the first step of this study; the notes are treated just like different words (E4, C4 …). Secondly, the order of the notes in the songs is used to define the context that will lead to the embeddings, as explained above. But the model does not take the order into consideration once the context has been defined. The sequence E4 C4 E4 for instance, and the sequence C4 E4 E4 will lead with a context size of 2 to the same context. I will come back to that feature of the study in the discussion.

## 2.5 Data analysis method

First, I examine the distribution of notes in the two datasets. Given the low dimensionality of the semantic space, the embeddings of the notes can be directly represented on a plane without any loss of information. This visualization provides an initial way to interpret the results, where geometrical properties may reflect semantic relationships between the notes.

To further analyze the results, I perform a multi-dimensional regression of the pitches (as initially coded in the data) on the embeddings (2-dimensional vectors). A high multiple correlation would indicate a strong link between the coordinates of the notes on the plane and their pitches: they could be reconstructed (or at least approximated) by an affine transformation of those coordinates.

To assess how the results depend on the order of the notes in the melody, I scrambled them in various ways. First, I generated a random melody with 16 uniformly distributed notes, randomly organized in a sequence, which I will refer to as the Baseline dataset. To ensure comparability, the training data has the same dimensions as those used for the 20 children's songs. Second, I created a melody from the 20 studied tunes by aggregating all the songs and randomly permuting the notes, resulting in a new tune—referred to as the Random song—with the same note distribution as the original 20 tunes but with a different sequence of notes. Third, I randomized the notes within each song separately, creating 20 Fake melodies, each with the same note distribution as its counterpart. Finally, I created a Random sonata in a similar manner.

Multiple correlations for all simulations were averaged over 100 runs of the model with different starting weights of the connection matrices. For the random variations of the datasets, I also modified the permutations for each run.

## 3. Results

### 3.1 The 20 children's songs

The 20 children's songs requires a vocabulary of 16 notes, whose frequency is given below.

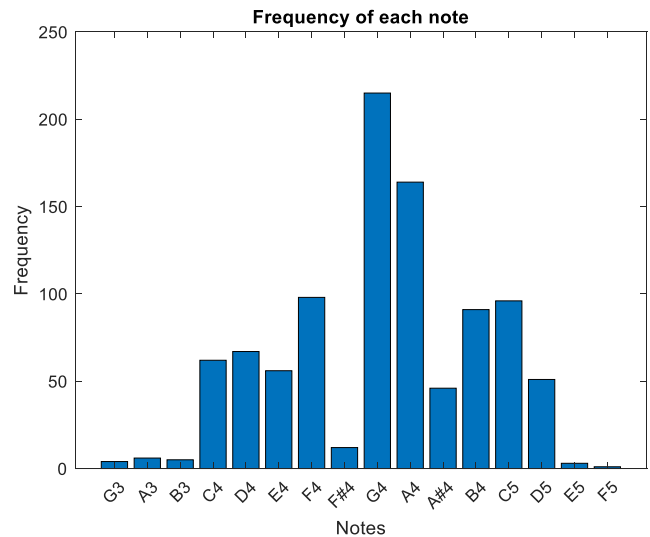

Figure 1. Distribution of the notes encountered in the 20 children's songs

After training, the embeddings of the 16 different notes, as the embedding size is two, can naturally be represented as points in a two-dimensional space. The following figure has been obtained with a context size of 4, for one of the simulations.

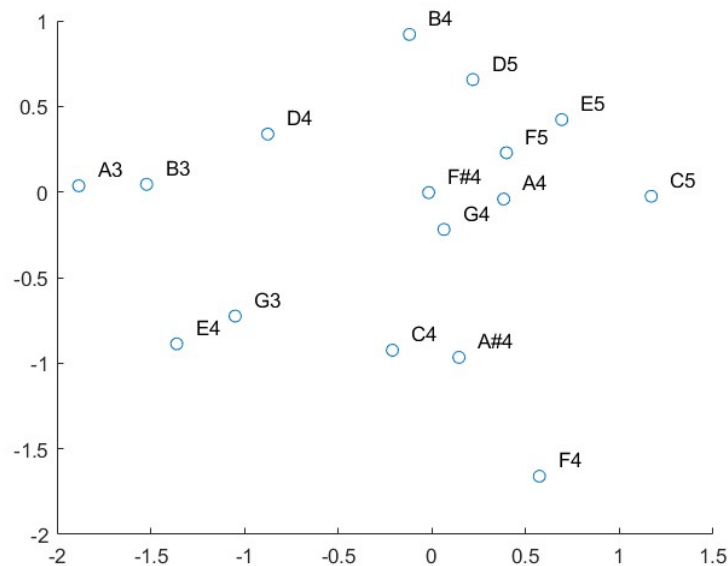

Figure 2. Embeddings of the notes encountered in the 20 children's songs after training for a context size of 4

The accuracy of the model, which generated the embeddings given above, achieved with a context size of 4, is 0.2 (measured on the validation data). This means that about 20% of the prediction made by the model were correct. But in embedding models, this measure is not very relevant. Notice that a validation measure of 1 would correspond to very monotonous music, where all notes can be easily anticipated.

The average multiple correlation between these vectors and the corresponding pitches is 0.86, for a context size of 4, as shown by a multiple regression analysis. I have replicated the analysis with different context sizes and the results are given below.

| Context size | 1 | 2 | 3 | 4 |
|---|---|---|---|---|
| Multiple correlation | 0.67 (SD:0.03) | 0.79 (SD:0.01) | 0.84 (SD:0.01) | 0.86 (SD:0.02) |

Table 1. Multiple correlation coefficients of the embeddings with the pitches for different context sizes for the 20 children's songs. The correlation coefficients are averaged over 100 runs of the model with different starting weights of the connection matrices. Their standard deviations are given into parenthesis.

The correlation coefficient increases as you consider more preceding notes to predict the upcoming one.

To graphically illustrate this result, a 3D plot of the regression plane obtained from a simulation with a context size of 4 and a multiple correlation of 0.87 is provided below.

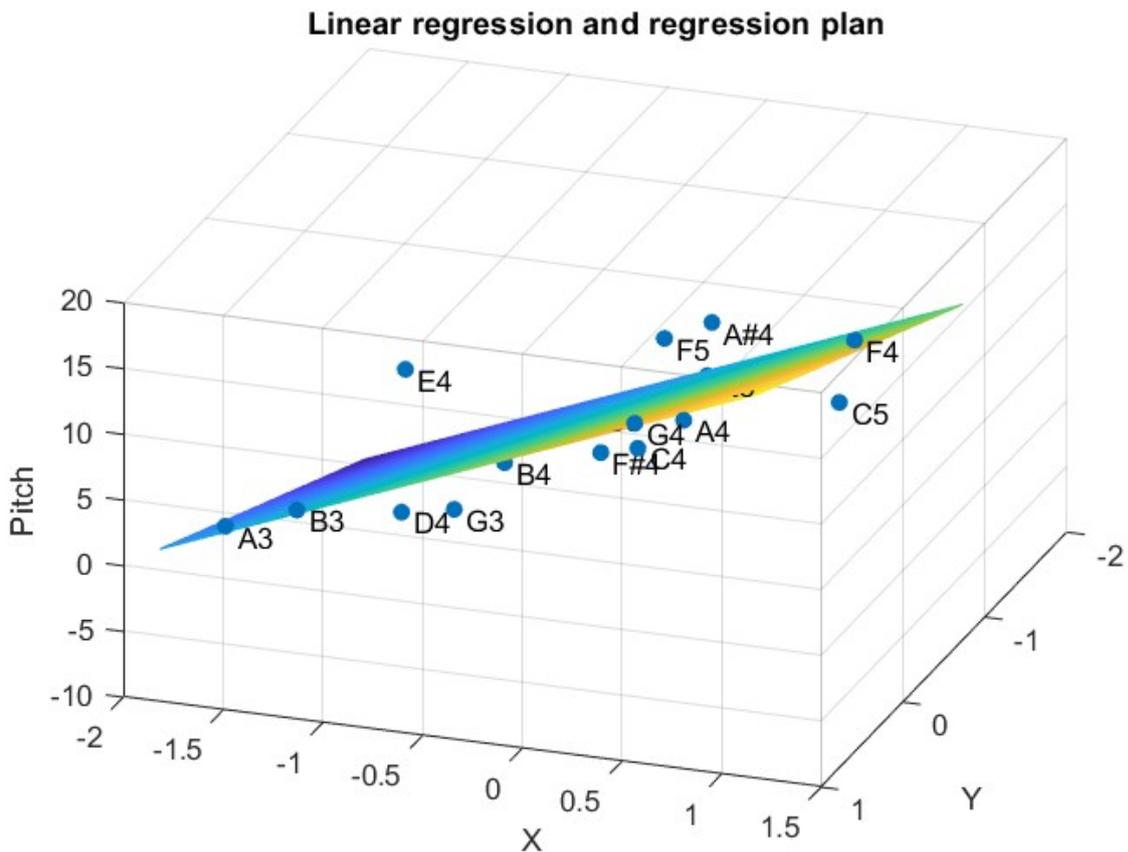

Figure 3. Regression of the pitches of the notes encountered in the 20 children's songs on their embeddings (X,Y) with an embedding size of 2 and a context size of 4

This figure demonstrates that an affine transformation of the embeddings, represented by a plane in the graphic above, provides a good approximation of the pitches of the notes.

## 3.2 The random variations of the children's songs

The results suggest that the order of notes in the studied melodies provides significant information about their pitch. However, these results might also be influenced by other characteristics of the melodies, such as the distribution of their notes, or the high number of note repetitions in the songs. To validate this interpretation, I conducted, as described in the methodological section, additional simulations using the same parameters. The table below presents the average multiple correlation coefficients obtained from these simulations.

| Context size | 1 | 2 | 3 | 4 |
|---|---|---|---|---|
| Baseline | 0.33 (SD:0.16) | 0.33 (SD:0.16) | 0.35 (SD:0.15) | 0.33 (SD:0.14) |
| Random song | 0.21 (SD:0.11) | 0.24 (SD:0.12) | 0.28 (SD:0.13) | 0.26 (SD:0.14) |
| Fake melodies | 0.76 (SD:0.03) | 0.79 (SD:0.04) | 0.79 (SD:0.04) | 0.78 (SD:0.05) |

Table 2. Multiple correlation coefficients of the embeddings with the pitches for different context sizes for the three different random variations of the children's songs. The correlation coefficients are averaged over 100 runs of the model with different starting weights of the connection matrices and different random permutations of the notes. Their standard deviations are given into parenthesis.

The differences with the results observed in the 20 tunes warrant several comments. The first two random datasets (Baseline and Random song) show correlation coefficients that are substantially lower than those obtained from the original children's songs, indicating no real possibility of reconstructing the pitches from the embeddings in a linear way. For instance, with a multiple correlation coefficient of 0.30, less than 10% of the variance of the pitches can be explained by the variations observed in the vector space. For a correlation coefficient of 0.80, this percentage goes up to 64%.

However, the results for the Fake melodies are more surprising. The average multiple correlation coefficients remain close to those obtained from the original tunes. This is likely due to specific features of the note distribution within each song: a limited number of notes, numerous repetitions, and the high frequency of certain notes. Consequently, random permutations do not fundamentally alter the neighborhood relationships. The Fake melodies still sound like children's songs.

The standard deviations of the multiple correlation coefficients are higher than those obtained from the children's songs. In this section, the random variations examined involve both the initial weights of the connection matrices and random permutations of the notes, which are varied across 100 different simulations. This means two independent sources of variations and consequently higher standard deviations.

## 3.3 The Bach Sonata

Regardless of the importance of the order of notes on the embeddings, it is also worth considering whether the results obtained are specific to very simple melodies, such as those found in nursery rhymes, or if comparable results can be observed in more sophisticated music. To address this question, I conducted, as already indicated, a simulation on the second dataset where the nursery rhymes are replaced by a Bach sonata. This sonata uses 25 different notes, which should make the multiple regression more convincing. They are distributed in the following way.

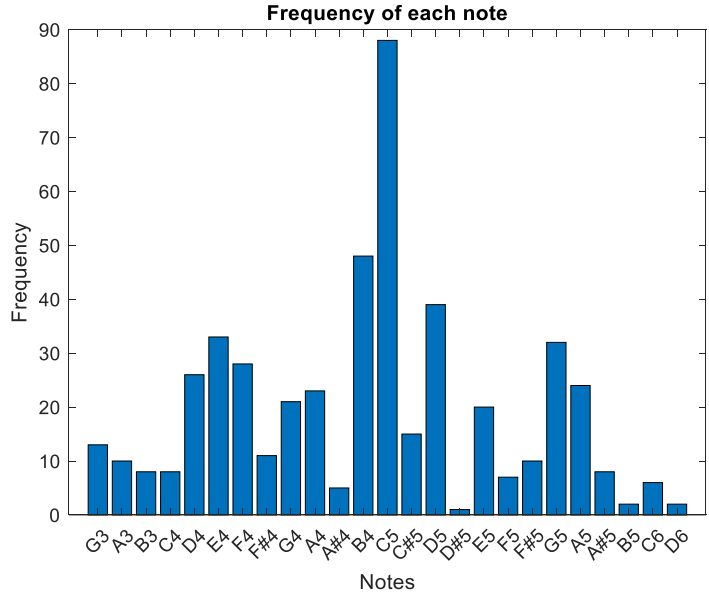

Figure 4. Distribution of the notes encountered in the Bach Sonata

The embeddings for a context size 2 are given below.

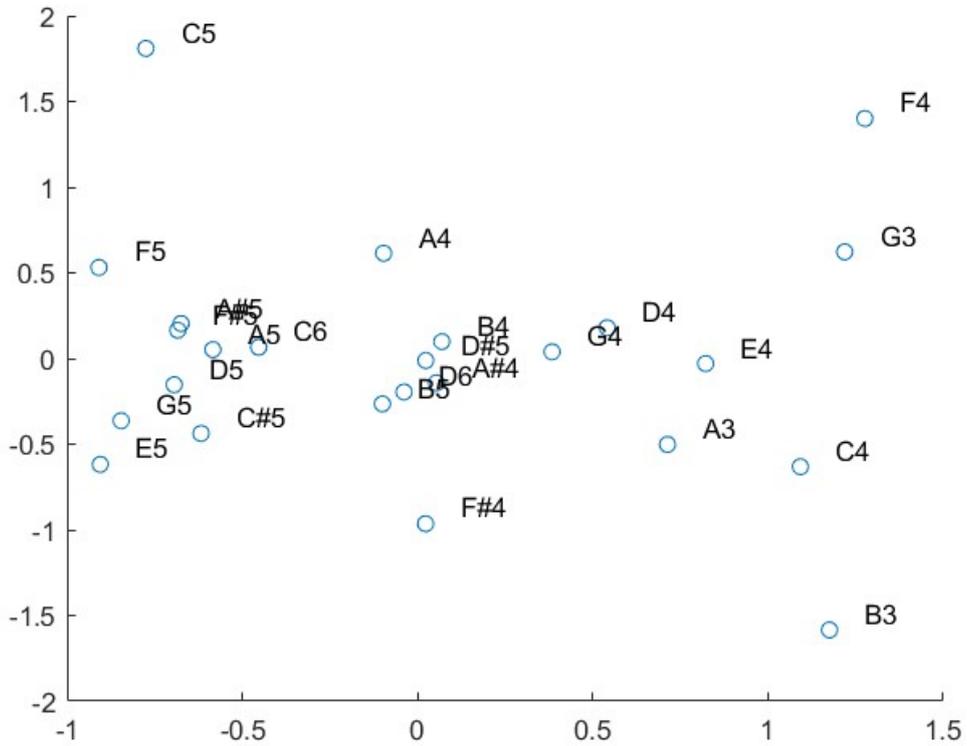

Figure 5. Embeddings of the notes encountered in the Bach Sonata after training for a context size of 2

The multiple correlation coefficients obtained with the regression analyses of the pitches on those embeddings for different context sizes are given below.

| Context size | 1 | 2 | 3 | 4 |
|---|---|---|---|---|
| Multiple correlation coefficient | 0.21 (SD:0.13) | 0.77 (SD:0.03) | 0.79 (SD:0.03) | 0.74 (SD:0.04) |

Table 3. Multiple correlation coefficients of the embeddings with the pitches for different context sizes for the Bach Sonata. The correlation coefficients are averaged over 100 runs of the model with different starting weights of the connection matrices. Their standard deviations are given into parenthesis.

The low value of the multiple correlation coefficient when only one note is taken into consideration (context size=1) is probably linked to the "sophistication" of the Sonata. Attempting to predict a note from the preceding one does not make much sense.

As for the 20 melodies, to graphically illustrate this result, a 3D plot of the regression plane obtained from a simulation with a context size of 2 and a multiple correlation of 0.79 is provided below.

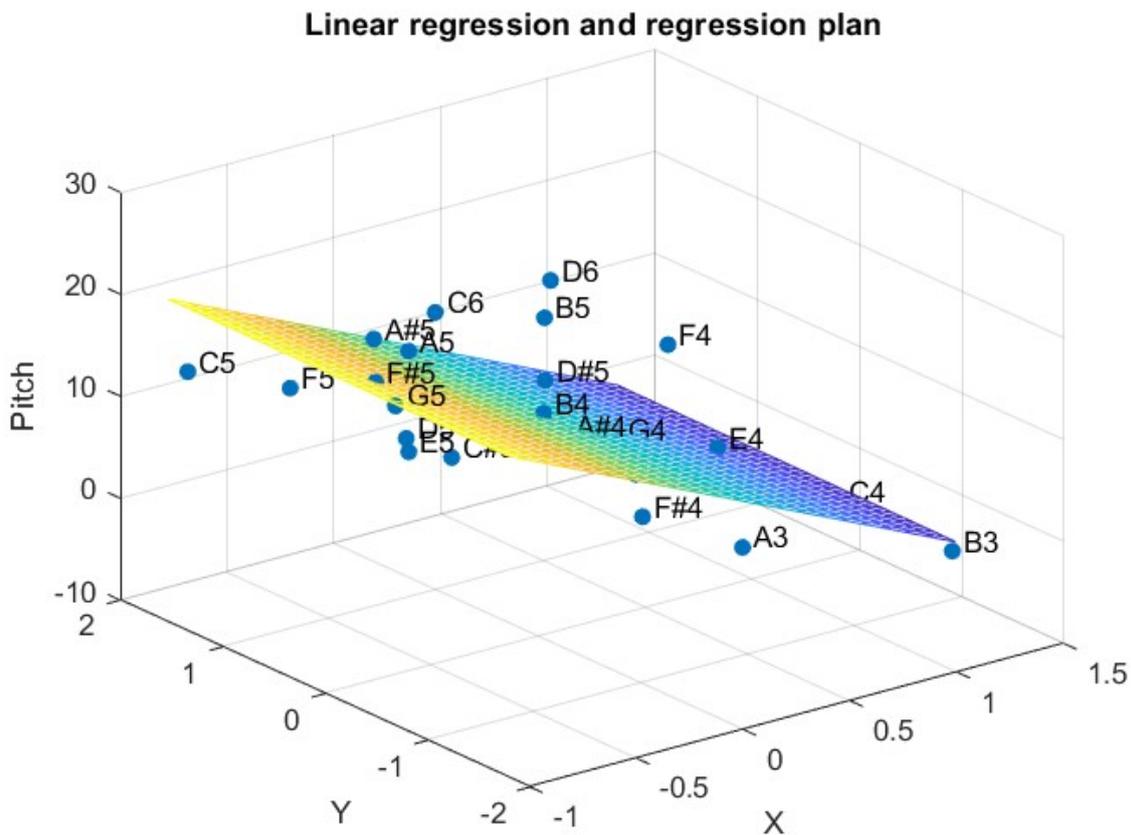

Figure 6. Regression of the pitches of the notes encountered in the Bach Sonata on their embeddings (X,Y) with an embedding size of 2 and a context size of 2

### 3.4 The Random Bach Sonata

The table below gives the values of the average multiple correlation coefficients obtained for the Random Bach Sonata.

| Context size | 1 | 2 | 3 | 4 |
|---|---|---|---|---|
| Multiple correlation coefficient | 0.18 (SD:0.10) | 0.24 (SD:0.11) | 0.21 (SD:0.11) | 0.21 (SD:0.11) |

Table 4. Multiple correlation coefficients of the embeddings with the pitches for different context sizes for the random Bach Sonata. The correlation coefficients are averaged over 100 runs of the model with different starting weights of the connection matrices and different random permutations of the notes. Their standard deviations are given into parenthesis.

Once again, the observed values are substantially lower than those calculated for the original Bach Sonata. This finding reinforces the results observed in the children's songs.

### 4. Discussion

The representation of musical notes in a semantic space by an embedding seems meaningful and leads to interesting results. Considering a musical piece as a simple text, whose words are notes, enables to associate numerical vectors to the notes, with a very simple network architecture. The vectors obtained in this way - the embeddings - are reputed to represent part of the meaning of the corresponding words. They convey information about semantic proximity, grammatical relations, functional relationships… In music, this seems to translate to information not only on tonal proximities and octaves, but also on the pitches. The results, obtained on what could be called a microcosm of tonal music, are promising. The way notes are chained is related to their pitch - this is not new, of course - but also contains "enough" information to retrieve these pitches. Therefore, in the melodies, at least those chosen in this study, there is a certain form of redundancy. What the note pitches tell us is already inscribed in their succession. Is this an explanation for the easy access to this kind of music?

The results I have established differ from those previously reported, to the best of my knowledge. For example, Garcia-Valencia, whose approach is somewhat similar to mine, used a much larger semantic space. This may have diluted the information on the pitches of the sounds. Additionally, the classification techniques he applied do not allow for analysis beyond identifying proximity relations.

One could question the size of the embedding space. Why not an embedding size of 1 or of 3? Additional simulations show that with an embedding size of 1, the correlation with the pitches is slightly higher for the 20 melodies with a context size of 2 (0.82) but is degraded for the Sonata (0.67). An embedding size of 3 does not improve the results.

The input data fed into the model are multi-hot vectors, which provide a particularly simple way to encode the data. A more traditional encoding method, where each note defining the context is represented by a one-hot vector, leads to very similar results. For example, with a context size of 2, the mean multiple correlations and their standard deviations are 0.81 (SD: 0.02), 0.26 (SD: 0.13), and 0.78 (SD: 0.03) for the 20 children's songs, the Random song, and the Sonata, respectively. These values are very close to the results obtained with multi-hot encoding (see Tables 1, 2, and 3). There are two main differences between the two types of encoding. First, multi-hot encoding does not differentiate the order of the notes within the context. However, this order can seem important: the sequence C E G is different from the sequence E C G. The study suggests (as is also true in language processing with the CBOW approach) that the concept of context is more relevant than the concept of succession for obtaining a semantic representation. In fact, a multi-hot representation of the context emphasizes chords, as demonstrated in the paper by Madjiheurem, Qu, and Walder (2016). Second, if notes are repeated in a given context (for instance, {C, C} to predict G in the sequence C C G), the note (here C) appears only once with multi-hot encoding.

## 5. Conclusions

Effectively predicting a note in a melody, based on the preceding ones, seems to require the construction of a "meaning" for the notes, which in our study takes the form of their pitch. To successfully complete its task, the network must construct a sufficiently general representation of the note in its hidden layer using a very small-dimensional vector. This involves extracting characteristics from the note that are useful for predicting what will happen next. In this context, pitch emerges as an essential aspect for making effective predictions.

## 6. Future work

Commenting on how the vectorial representation of notes varies with context size (e.g., the Bach Sonata with a context size of 1), embedding size, and on note representation goes beyond the scope of this study. However, further exploration could help characterize pieces of music from the configuration of points in vectorial spaces.

The analysis performed on the first 10 children's songs, with a context size of 2, yields an average multiple correlation coefficient of 0.70. This suggests that the link with pitches only becomes apparent when the input data size is sufficiently large.

It would also be interesting to extend this study by examining the representation of intervals between notes, as done in numerous studies on the subject (Garcia-Valencia, 2020; Defays, French, and Tillmann, 2023). However, calculating these intervals requires a certain form of quantification of perceived elementary sounds, which justifies the approach chosen in this study.

The results obtained by Defays, French, and Tillmann (2023) were based on a simplified version of existing melodies, considering only the melody track and not the harmonic context. Viewing music as a set of words allows for the incorporation of more complex musical information, as demonstrated by Hongru Liang, Wenqiang Lei, Paul Yaozhu Chan, Zhenglu Yang, Maosong Sun and Tat-Seng Chua (2020), including rhythm, dynamics, harmonic context, and possibly timbre. However, this approach will require more complex models.